\documentclass[letterpaper]{article} 
\usepackage{aaai25}  
\usepackage{times}  
\usepackage{helvet}  
\usepackage{nameref}
\usepackage[skins]{tcolorbox}
\usepackage{courier}  
\usepackage{adjustbox}
\usepackage[hyphens]{url}  
\usepackage{graphicx} 
\usepackage{longtable}
\usepackage{natbib}  
\usepackage{caption} 
\usepackage{multicol}
\usepackage{enumitem}   %
\usepackage[table]{xcolor}  
\definecolor{lightgray}{gray}{0.9}  
\usepackage{booktabs}
\usepackage{algorithm}
\usepackage{algorithmic}

\usepackage{newfloat}
\usepackage{listings}
\DeclareCaptionStyle{ruled}{labelfont=normalfont,labelsep=colon,strut=off} 
\floatstyle{ruled}
\newfloat{listing}{tb}{lst}{}
\floatname{listing}{Listing}
\title{DualAlign: Generating Clinically Grounded Synthetic Data}

\affiliations{
 Rumeng Li,
 Umass Amherst,
 VA Bedford Healthcare System,

 Xun Wang,
 Micrsoft,

 Hong Yu,
 Umass Lowell,
 Umass Amherst,
 VA Bedford Healthcare System
 
}

\usepackage{bibentry}

\begin{document}

\maketitle

\begin{abstract}
Synthetic clinical data are increasingly important for advancing AI in healthcare, given strict privacy constraints on real-world EHRs, limited availability of annotated rare-condition data, and systemic biases in observational datasets. While large language models (LLMs) can generate fluent clinical text, producing synthetic data that is both realistic and clinically meaningful remains challenging. We introduce DualAlign, a framework that enhances statistical fidelity and clinical plausibility through dual alignment: (1) statistical alignment, which conditions generation on patient demographics and risk factors; and (2) semantic alignment, which incorporates real-world symptom trajectories to guide content generation. Using Alzheimer’s disease (AD) as a case study, DualAlign produces context-grounded symptom-level sentences that better reflect real-world clinical documentation. Fine-tuning an LLaMA 3.1–8B model with a combination of DualAlign-generated and human-annotated data yields substantial performance gains over models trained on gold data alone or unguided synthetic baselines. While DualAlign does not fully capture longitudinal complexity, it offers a practical approach for generating clinically grounded, privacy-preserving synthetic data to support low-resource clinical text analysis.
\end{abstract}

%
\section{Introduction}
Synthetic data generation has emerged as a critical strategy for advancing machine learning in the clinical domain \cite{chen2021synthetic, rujas2025synthetic, giuffre2023harnessing}, where access to real-world patient records is tightly constrained by privacy regulations, institutional policies, and compliance requirements \cite{liu2022real,rieke2020future}. By producing privacy-preserving datasets that retain the clinical characteristics of real health records, synthetic data enables broader research collaboration, reproducibility, and robust model development. Recent advances in large language models (LLMs) have demonstrated promise in generating clinical text \cite{smolyak2024large, barr2025large}, but existing approaches are largely general-purpose and lack the domain-specific conditioning needed to generate contextually rich and clinically diverse content reflective of particular diseases.
This limitation is especially critical for chronic conditions characterized by slow, multifaceted progression across interdependent clinical domains—such as neurodegenerative, cardiovascular, or metabolic disorders—where accurate representation of temporal dynamics and domain interactions is essential for meaningful downstream modeling. 

Alzheimer’s disease (AD) exemplifies this class of conditions, combining prolonged preclinical phases with progressive, system-wide deterioration. Its progressive nature involves coordinated decline across cognitive, functional, neuropsychiatric, and physiological domains, often unfolding over decades before formal diagnosis \cite{van2023towards, liss2021practical}. Effective early detection and disease prediction modeling thus rely on data that capture these nuanced, interwoven progression patterns. However, such real-world data are severely restricted due to privacy concerns, precisely where synthetic data could play a transformative role. 

Despite growing interest in synthetic health data, prior work on AD-specific synthetic clinical text remains limited. Few existing efforts have addressed the challenge of generating richly contextualized narratives tailored to the longitudinal and multifactorial nature of AD. Current synthetic data generators fall short in this domain \cite{li2023two,loni2025review}: they fail to preserve temporal coherence, underrepresent domain-specific correlations, and neglect key demographic modifiers such as age, sex etc. As a result, generated narratives are often fragmented and sometimes clinically implausible, undermining the robustness of downstream analysis models. 

To address these challenges, we argue for a new paradigm in synthetic clinical data generation, one that is disease-aware and consistent by design. In this work, we focus on AD as a critical use case, proposing a framework that integrates disease-specific priors into LLM-based generation to produce synthetic patient trajectories with high clinical fidelity and temporal plausibility.


We introduce the DualAlign, a novel framework for generating synthetic clinical data tailored to clinical data generation. DualAlign employs a dual alignment strategy to bridge statistical fidelity and clinical plausibility through two complementary mechanisms: (1) persona-driven alignment, which conditions narrative generation on patient demographics and risk factors to produce realistic patient profiles aligned with population patterns; and (2) longitudinal symptom alignment, which uses real-world symptom trends to guide the model in generating narratives enriched with a diverse and comprehensive set of AD signs and symptoms. This strategy promotes not only realistic symptom mentions but also greater diversity and contextual grounding, addressing key limitations of prior synthetic data approaches that produce fragmented or overly generic outputs.

From these generated narratives, we derive contextually grounded and diverse sentences describing AD signs and symptoms, automatically annotated using an LLM-based annotator guided by human-curated protocols. This yields a privacy-preserving, richly varied dataset that fills a critical gap in existing resources, which often lack both narrative context and symptom diversity.

While generating fully realistic longitudinal EHR narratives remains a complex and open challenge, DualAlign provides a first step by producing synthetic narratives that span multiple years, albeit in a compressed form. These narratives serve as a valuable scaffold for generating context-aware symptom data, offering opportunities for future research.

Our contributions are threefold:
\begin{itemize}
\item We introduce DualAlign, a dual alignment framework that integrates persona-driven simulation with longitudinal symptom alignment to generate synthetic clinical narratives enriched with diverse and contextually grounded AD signs and symptoms.

\item We construct and publicly release the first synthetic, privacy-preserving dataset of LLM-annotated AD sign and symptom mentions, derived from DualAlign-generated narratives and annotated using human-curated protocols. Classification experiments show that augmenting gold data with DualAlign improves model performance over gold-only and unguided synthetic baselines. Even in isolation, DualAlign achieves moderate accuracy on both binary (0.82 vs. 0.87 gold-only) and multi-class (0.53 vs. 0.70 gold-only) tasks, supporting its utility as a standalone training resource when real-world annotations are limited.

\item We provide a preliminary exploration of synthetic longitudinal EHR narrative generation, identifying key challenges and laying the groundwork for future improvements in modeling clinical complexity in synthetic data.
\end{itemize}

\section{Related Work}

\paragraph{Synthetic Data Generation in Healthcare}
Synthetic data generation has become a key strategy for enabling machine learning in clinical settings, where access to real-world patient data is limited by privacy regulations and institutional barriers. Early work primarily focused on structured data synthesis using GANs and VAEs to simulate patient demographics, diagnoses, and lab values while retaining core statistical properties \cite{chen2021synthetic, liu2025preserving, rujas2025synthetic, hernandez2022synthetic, gonzales2023synthetic}. While effective for benchmarking, these methods do not extend easily to free-text narratives, which require preservation of semantic nuance and temporal coherence.

\paragraph{LLMs for Clinical Text}
Recent advances in LLMs such as GPT-3 and GPT-4 have enabled the generation of synthetic clinical text, including discharge summaries, radiology reports, and SOAP-style notes \cite{taloni2025exploring, peng2023study, mawaldi2024synthetic, ganzinger2025automated, williams2025physician}. While these models produce fluent and stylistically appropriate outputs, most current approaches operate at the sentence or passage level and lack temporal grounding. Moreover, without domain-specific conditioning or fine-tuning, LLMs are prone to hallucinations—generating implausible or factually incorrect content \cite{kim2025medical, shah2024accuracy}—limiting their reliability for downstream applications such as risk modeling or early detection.

\paragraph{Synthetic Data for AD Research} AD, the leading cause of dementia, presents unique challenges for synthetic data generation due to its gradual, multifaceted progression across clinical domains. While machine learning models have been developed for AD risk stratification and phenotyping using EHR data \cite{li2023early, xu2020data, wang2021development, tjandra2020cohort}, synthetic text efforts remain limited. Existing datasets typically rely on structured simulations or rule-based generation \cite{li2023two, muniz2021virtual, sajjad2021deep}, lacking the narrative diversity and contextual realism found in real clinical notes.

These limitations underscore the need for synthetic methods that better capture clinical context and symptom plausibility—especially for complex conditions like AD.

\section{Method}
DualAlign is a synthetic data generation framework that produces demographically representative, clinically plausible, and symptom-rich narratives of AD. It comprises three components: (1) real-world pattern extraction, (2) language model–based generation guided by structured prompts, and (3) automated symptom annotation. Together, these modules enable the creation of diverse, context-aware datasets for privacy-preserving clinical NLP research. Here we use GPT-4 \cite{achiam2023gpt} as the LLM engine.

\paragraph{Step 1: Extracting Key Statistics and Clinical Patterns from Real-World Data}
We begin by analyzing longitudinal records from the U.S. Department of Veterans Affairs (VA) to extract statistics that inform synthetic patient personas and symptom trajectories. While clinically detailed, the VA dataset is demographically skewed—predominantly older, male, and white. To enhance representativeness, we supplement this analysis with epidemiological data from national AD reports \cite{better2023alzheimer}, which provide population-level distributions across age, sex, and race/ethnicity.

In parallel, we quantify the occurrence of AD signs and symptoms across  subgroups in the VA cohort to model their typical emergence and progression. These trends inform both the demographic configuration and longitudinal symptom structure of our synthetic narratives.

\paragraph{Step 2: Generating Data with Statistical and Semantical Guidance}
Using the extracted statistics, we simulate synthetic patient personas stratified by demographics and AD risk factors, ensuring stronger representation of underrepresented populations. For each patient, we generate multi-year clinical notes by sampling visit types and inserting symptom-related keywords in contexts. These structured attributes—demographics, visit type, temporal context, and symptoms—are passed as prompts to an LLM to produce realistic and diverse clinical narratives.

\paragraph{Step 3: Annotating Generated Data for Downstream Tasks}
Finally, we apply an LLM-based annotator guided by human-curated protocols to extract and label symptom-relevant sentences across five clinically validated categories. This process yields a high-quality, privacy-preserving dataset suitable for downstream tasks such as early AD risk prediction and symptom extraction.

The DualAlign generated data are then tested on real-world data to verify their usefulness.

\begin{figure*}
    \centering
    \includegraphics[width=1\linewidth]{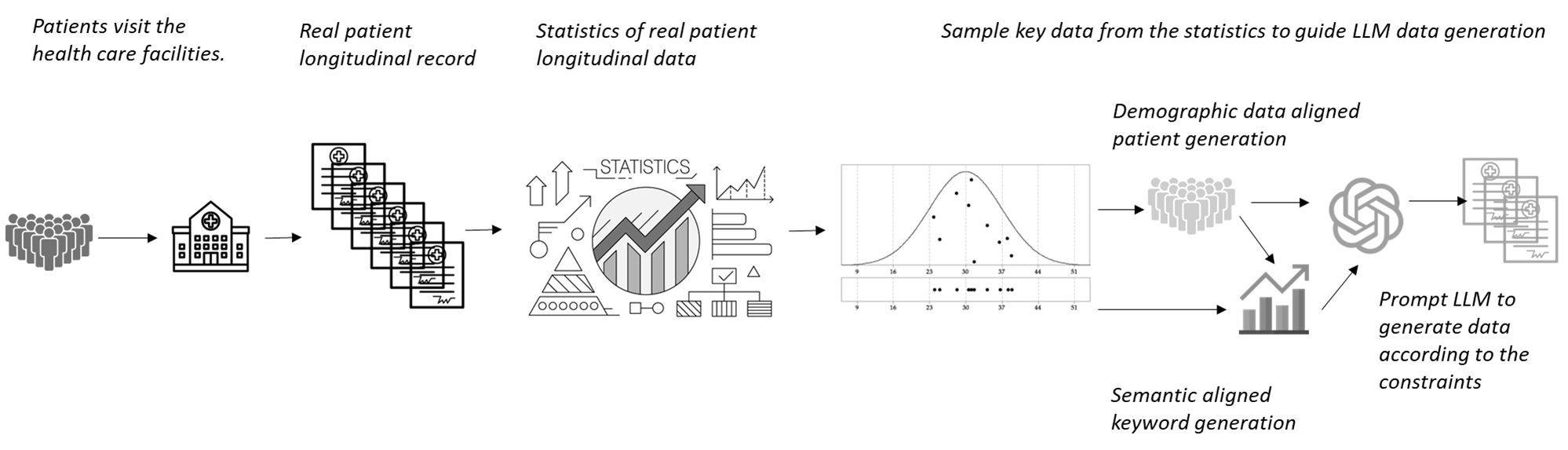}
    \caption{DualAlign data generation workflow}
    \label{fig:datagen}
\end{figure*}

The overall workflow is summarized in Figure~\ref{fig:datagen}, and we present each component of the methodology in the subsequent section.

\subsection{AD Patient Cohort Building}

To guide synthetic data generation with real-world patterns, we analyzed EHR data from the VA Corporate Data Warehouse (CDW), one of the largest integrated healthcare databases in the U.S. The VHA provides comprehensive medical services to veterans across over 1,200 healthcare facilities. Its EHR system contains extensive patient data, including demographics, diagnoses, clinical notes, and visit records.

We constructed an AD case cohort by identifying patients with at least two separate AD diagnoses between October 1, 2015 (the implementation date of ICD-10 codes), and September 30, 2022 (the end of our study period). To guarantee expert-level assessment, at least one of these diagnoses must have been made at a specialty clinic—such as neurology, geriatrics, geriatric patient aligned care team (GeriPACT), mental health, psychology, psychiatry, or geriatric psychiatry—by a provider specializing in neurology, vascular neurology, psychiatry, neuropsychology, or geriatric medicine. This resulted in a cohort of 35,308 AD cases, with a mean (SD) age at diagnosis of 79.0 (8.4) years. The cohort was demographically skewed: only 2.0\% were women, and the majority were White (78.9\%), with 12.9\% Black or African American, and 6.6\% Hispanic or Latino.

We extracted longitudinal clinical notes for these patients spanning the period from 2000 to 2022, focusing on note types relevant to AD diagnosis and management, including notes generated from primary care visits, emergency visits, home-based primary care (HBPC), memory clinics, neurology, neuropsychology, geriatrics, psychiatry, psychology, cognitive care nursing, mental health clinics, compensation and pension examinations, and consultation visits. This real-world cohort served as a reference for both patient demographics and symptom documentation trends. The study was approved by the Institutional Review Board (IRB) at the VHA Bedford Healthcare System.

\subsection{Patient Profile Generation: Demographics and Risk Factors}

While the VHA AD cohort provides valuable longitudinal data, its demographic composition is not representative of the broader AD population—particularly underrepresenting women and several racial and ethnic groups due to the characteristics of the veteran population \cite{AHRQ2020Vets}. This demographic skew contrasts with epidemiological evidence showing that women have a higher lifetime risk of AD and that prevalence varies significantly across racial and ethnic groups.
To address this, we explicitly align our synthetic cohort’s demographic distribution with national AD prevalence estimates from the 2024 Alzheimer’s Disease Facts and Figures report \cite{better2023alzheimer}. Age, sex, and race/ethnicity attributes are sampled using stratified probability distributions to reflect population-level statistics, ensuring representativeness across key demographic strata.

Beyond demographics, we incorporate approximately 50 established AD risk factors and social determinants of health (SDOH), derived from prior epidemiological and clinical studies \cite{livingston2024dementia,jones2025environmental,rohr2022social,stites2022establishing}. These include genetic predispositions (e.g., family history of AD), common comorbidities (e.g., hypertension, diabetes, cardiovascular disease, stroke history), neuropsychiatric and inflammatory conditions (e.g., depression, anxiety, chronic inflammation), and social/environmental factors such as housing instability. The complete list of AD risk factors and SDOH together with the prevalence is provided in Appendix~\nameref{sec:ad-risk-factors}. Risk factor values are sampled using categorical distributions.

This structured, probabilistically grounded approach generates a diverse and epidemiologically calibrated set of synthetic patient personas. These personas serve as the foundation for LLM-driven clinical narrative generation, ensuring that downstream synthetic notes reflect both the complexity and heterogeneity of real-world AD trajectories.

\subsection{Temporal Alignment of Clinical Notes: Visit Frequency and Type}
After constructing the synthetic patient persona, we simulate a clinically plausible trajectory of note generation spanning the ten years prior to AD diagnosis. This involves aligning the number and types of clinical notes per year with empirical patterns observed in the VA cohort (Table~\ref{tab:note_types}, Appendix). We also map the time to AD stage (see Appendix \nameref{sec:temporalcontext}) to provide stage information as LLM are not senstive to numbers/time.
Empirical bootstrapping is used unless otherwise specified.

\paragraph{Modeling Visit Frequencies.}
We simulate visit frequencies by sampling from time-stratified empirical distributions derived from real-world utilization patterns. These reflect a gradual increase in healthcare engagement as cognitive impairment progresses, with primary care remaining the dominant visit type across all windows. Specialty visits—including neurology, memory clinics, neuropsychology, and psychiatry—become increasingly prevalent in the final years prior to diagnosis, aligning with clinical expectations of escalating diagnostic evaluation and management.

\paragraph{Sampling Visit Types.}
For each synthetic visit, a note type is drawn from a time-dependent distribution including: primary care, emergency visits, home-based primary care (HBPC), and specialty services such as neurology, geriatrics, psychiatry/mental health, neuropsychology, and memory clinics. Each note is explicitly anchored to a year relative to diagnosis and is structurally tied to the patient persona and care context.

\subsection{Semantic Alignment of Clinical Notes: Constraint via Clinician-Curated Keywords} \label{sec:semantic_alignment}
To enhance clinical validity, we guide narrative generation using a curated lexicon of AD-relevant signs and symptoms, aligning keyword sampling with real-world temporal and categorical trends observed in our created VA cohort.

\paragraph{Lexicon Construction.}
Based on an extensive literature review and prior expert-curated keyword efforts \cite{better2023alzheimer, livingston2024dementia, wang2021development}, we developed a lexicon of 122 AD-relevant keywords. Six domain experts—including clinicians and researchers in cognitive aging, neurology, and epidemiology—collaboratively proposed and validated candidate terms across six major symptom domains.
\begin{itemize}
    \item \textbf{Cognition–speech/language:} e.g., word-finding difficulty, aphasia, impaired comprehension
    \item \textbf{Cognition–memory:} e.g., memory loss, forgetfulness, difficulty recognizing people or places
    \item \textbf{Cognition–learning and perception:} e.g., impaired attention, visuospatial disorientation, executive dysfunction
    \item \textbf{Assistance needed:} e.g., difficulty with ADLs/iADLs such as dressing, medication adherence, or finances
    \item \textbf{Physiological changes:} e.g., gait instability, sleep disturbance, sensory deficits, swallowing issues
    \item \textbf{Neuropsychiatric symptoms:} e.g., depression, apathy, agitation, impaired insight, personality change
\end{itemize}
The full keyword list is provided in Appendix~\nameref{sec:keyword-list}.

\paragraph{Keyword Pattern Alignment.}
To simulate clinically realistic symptom emergence, we analyze keyword usage along two axes: (i) frequency per note across time windows, and (ii) relative prevalence by symptom category. We construct stratified sampling tables based on keyword distributions observed in our AD case cohort to guide generation. Corresponding statistics are summarized in Appendix Tables~\ref{tab:keyword_count} and~\ref{tab:keyword_proportion}.

In real-world EHRs, AD-relevant symptom mentions are sparse—averaging only 2.7 to 4.2 keywords per note in the years leading up to diagnosis. While this reflects typical documentation patterns, such sparsity limits the utility of the generated notes for training extraction and classification models. To address this, we apply a 5× multiplier to the real-world keyword frequency estimates. This design choice increases symptom mention density, enabling more efficient training while maintaining realistic category-level proportions.

To enforce semantic alignment, we first sample the number of keyword mentions based on the given trend, then we sample the category for each keyword mention based on the category distribution (categorical distribution), and finally select the specific keyword from the corresponding keyword set of the chosen category. Keywords are sampled uniformly from a categorical distribution.

Sampled keywords are embedded into LLM prompts, encouraging integration into SOAP-style clinical text. Prompt templates also introduce variability via abbreviations, domain-specific phrasing, and occasional typos to emulate natural documentation. 

\subsection{Prompt Template }

\begin{figure}[h]
    \centering
    \includegraphics[width=\linewidth]{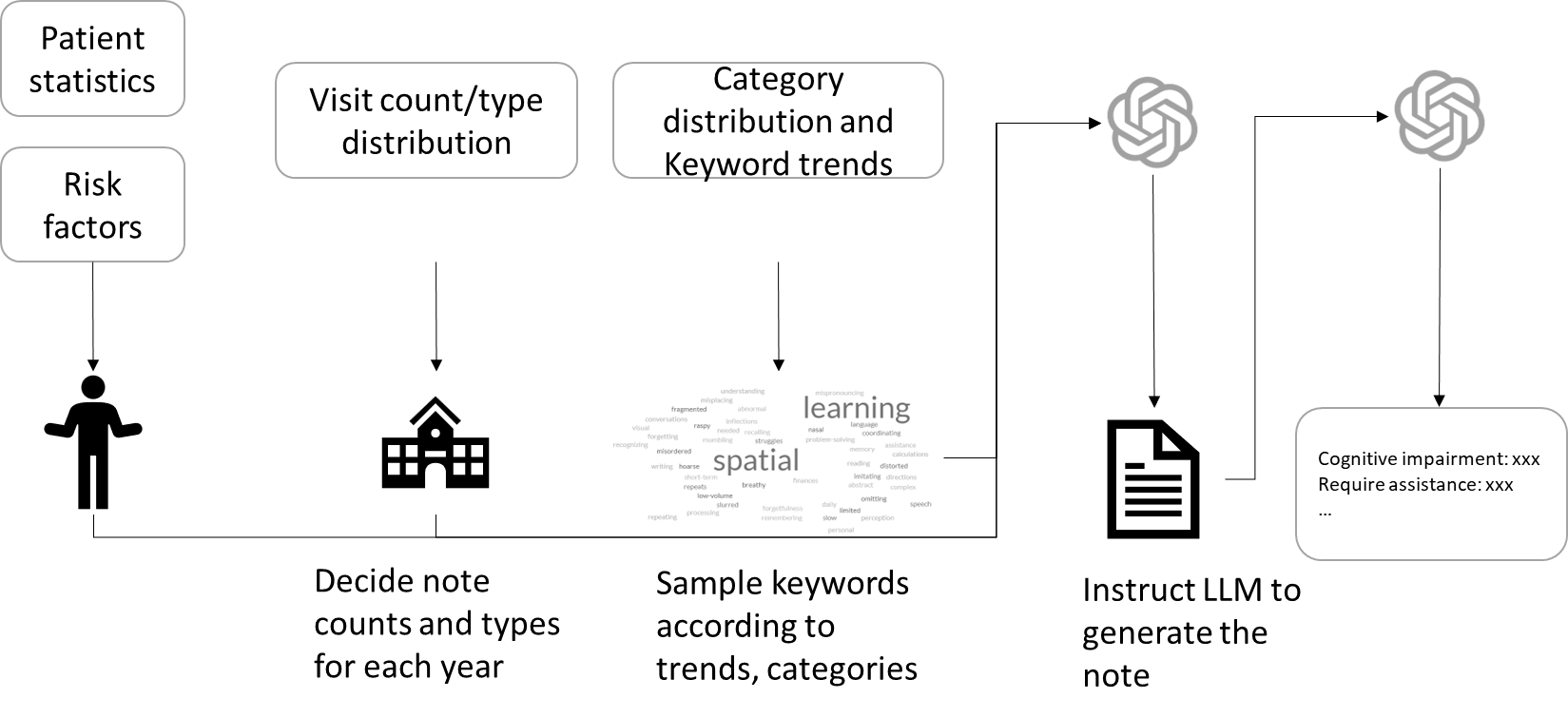}
    \caption{Construction of the detailed instructions for note generation.}
    \label{fig:prompt-framework}
\end{figure}

Figure~\ref{fig:prompt-framework} summarizes the overall prompt construction workflow, which integrates patient demographics, visit timing, and symptom trends into structured instructions. This pipeline enables controlled and contextually grounded note generation aligned with real-world AD trajectories.

\begin{figure}
    \centering
    \includegraphics[width=1\linewidth]{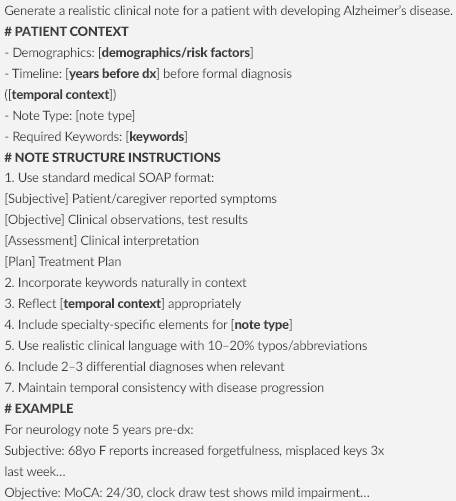}
    \caption{The prompt template to generate note content.}
    \label{fig:prompt}
\end{figure}

The final prompt template used in the DualAlign is shown in Figure~\ref{fig:prompt}. 
As illustrated, it embeds key clinical knowledge and progression patterns associated with AD, ensuring that the generated content remains clinically plausible and contextually aligned with the natural history of the disease. An example of a generated note is provided in Appendix~\nameref{sec:example-note}.


\subsection{Experiments on Downstream Task}

\paragraph{LLM Annotation}
To annotate AD symptoms from synthetic clinical notes, we employ an LLM-based annotator guided by structured, clinician-curated labeling protocols. These protocols are adapted from established guidelines introduced in prior work~\cite{li2023two}, which originally included nine categories such as diagnostic tests and formal cognitive assessments. In this study, we focus on five categories that emphasize observable signs and symptoms likely to appear in routine clinical documentation.

This symptom-focused taxonomy was intentionally narrowed by domain experts to capture subtle early indicators—such as forgetfulness, behavioral shifts, or emerging difficulties with daily function—that may precede formal diagnostic procedures and often go unnoticed in unstructured EHR narratives. By excluding categories like cognitive assessments and diagnostic tests (which typically follow physician suspicion), we aim to facilitate scalable annotation and downstream analysis of naturally occurring early-stage AD progression signals.

\begin{itemize}
\item \textbf{Cognitive impairment:} memory loss, confusion, word-finding difficulty
\item \textbf{Concerns raised by others:} caregiver-reported changes or observations
\item \textbf{ Require assistance / Functional impairment:} decline in ADLs/iADLs, supervision needs
\item \textbf{Physiological changes:} motor or sensory decline, sleep disturbance, incontinence
\item \textbf{Neuropsychiatric symptoms:} agitation, hallucinations, anxiety, apathy
\end{itemize}
Detailed annotation guidelines are available in Appendix~\nameref{sec:annotation-guideline}.
The LLM annotator operates at the sentence level, identifying relevant mentions across the longitudinal EHR notes generated by DualAlign. This process results in a high-coverage, contextually grounded synthetic dataset of 233,014 labeled sentences, capturing AD signs and symptoms purely through LLM generation and annotation.
\paragraph{Benchmark}
To evaluate DualAlign-generated data, we use a gold-standard dataset of 11,571 human-annotated sentences from 5,112 longitudinal notes of 76 real-world AD patients~\cite{li2023two}. These annotations were created under physician supervision using the same guidelines applied in the LLM annotation process described above.
(Cohen’s $\kappa$ = 0.868). We split this dataset into 80/10/10 training, validation, and test sets.

We fine tuned the Llama 3.1 8B model \cite{meta2024llama31} on the training set to classify sentences in EHR notes. Inputs were tokenized and truncated/padded to 128 tokens. Training used a batch size of 16 on each GPU with gradient accumulation (effective batch size 64), AdamW optimizer, linear warmup ratio(0.1) followed by decay. A classification head with dropout (0.1) was added to the first token’s representation. The model was trained for 3 epochs with early stopping based on validation macro F1, evaluated every 200 steps. Parameters are as follows: $batch\_size=64, lr=2e^{-5}, \beta_1=0.9, \beta_2=0.999,weight\_decay=1e^{-2}$.

\section{Results}
To assess the contribution of synthetic data—both alone and in combination with human-annotated data—we designed classification experiments with five training configurations:

\begin{itemize}
    \item \textbf{Gold Only:} Establishes the \textit{human-annotated baseline}, serving as a reference for performance without synthetic augmentation.

    \item \textbf{Gold + Bronze:} Prior synthetic augmentation baseline from existing work~\cite{li2023two}, combining the gold human-annotated dataset with a prompt-based LLM-generated synthetic dataset (GPT-4) that was directly annotated. This synthetic data lacks demographic and semantic constraints.

    \item \textbf{Gold + DualAlign (matched):} Size-controlled comparison to isolate the effect of improved realism and context in DualAlign.

    \item \textbf{Gold + DualAlign (full):} Upper bound using the full DualAlign dataset to assess large-scale gains.

    \item \textbf{DualAlign (full) Only:} Evaluates whether synthetic data alone can yield competitive classification performance.
\end{itemize}

We exclude the \textit{silver} dataset from~\cite{li2023two} as it involves LLM annotation of real-world EHRs (MIMIC-III) rather than full synthetic generation, making it methodologically incompatible with DualAlign.
Table~\ref{tab:category_stats} summarizes the sentence counts per symptom category across the gold, bronze, and DualAlign datasets used in these configurations.

Following prior work~\cite{li2023two}, we assessed performance on two tasks:

\begin{itemize}
    \item \textbf{Binary classification:} Detect whether a sentence describes an AD-related sign or symptom. Negative examples were sampled from the same 76-patient longitudinal corpus, maintaining a 5:1 negative-to-positive ratio.

  \item \textbf{Multi-class classification:} Assign each AD-related sentence to one of the five clinically defined symptom categories detailed in the methods section.
\end{itemize}

All configurations were evaluated on the same held-out gold-standard test set to ensure consistent comparison.

As shown in Table~\ref{tab:binary_perf_transposed}, augmenting the gold training set with DualAlign (full) yielded the strongest binary classification performance, achieving an F1 score of 0.84 and Accuracy of 0.95—substantially surpassing the gold-only baseline. Notably, Gold + DualAlign (matched) also outperformed Gold + Bronze, highlighting the value of demographic and symptom-guided alignment. Even using DualAlign (full) alone, without any gold data, achieved an Accuracy of 0.82, suggesting that well-designed synthetic data can provide meaningful signal in low-resource settings, despite a lower F1 of 0.47.

In the sentence-level AD signs/symptoms classification task (Table~\ref{tab:multiclass_f1}), we observe consistent performance gains across all six symptom categories when augmenting the gold-standard training set with DualAlign (full) data. Notably, the largest relative improvements occur in categories that are typically more challenging to annotate, such as \textit{Requires Assistance} (F1 from 0.57 to 0.72) and \textit{Concerns by Others} (F1 from 0.46 to 0.60). Substantial gains are also observed in \textit{Physiological Changes} and \textit{Neuropsychiatric Symptoms}, further highlighting the model’s ability to generalize across diverse symptom types. These findings suggest that DualAlign-generated data effectively enhances coverage of underrepresented and semantically nuanced categories, contributing to more robust multi-class classification performance.

\begin{figure}[t]
    \centering
    \includegraphics[width=\linewidth]{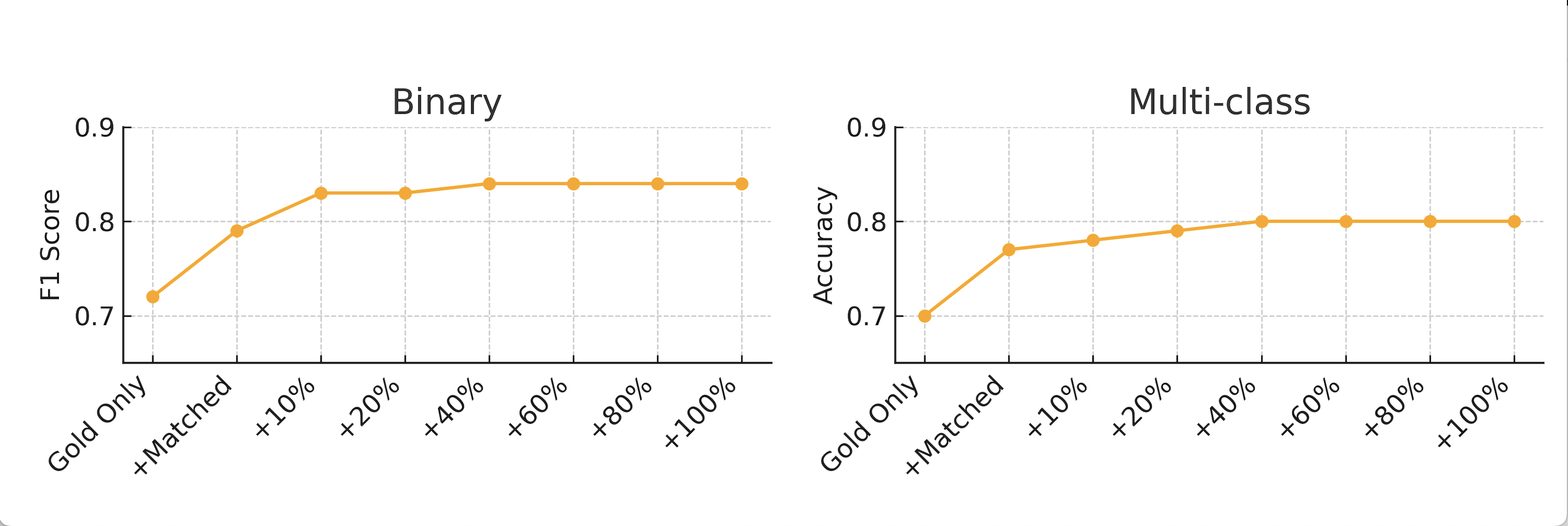}
    \caption{Performance with incremental addition of DualAlign-generated data. Left: Binary F1 score. Right: Multi-class accuracy. }
    \label{fig:increment}
\end{figure}

To examine the impact of synthetic data volume on performance, we incrementally added DualAlign-generated notes to the gold training set in 10\%/20\% increments. As shown in Figure~\ref{fig:increment}, both binary F1 score (left) and multi-class accuracy (right) improve substantially with additional synthetic data, plateauing around the 40\% mark. Beyond this threshold, further data yield diminishing returns. These results suggest that a moderate amount of well-aligned synthetic data is sufficient to approach peak performance, reducing the need for extensive manual annotations. The trend holds consistently across both binary and multi-class settings, highlighting the practical value of DualAlign in enhancing training data quality at scale.

Together, these findings suggest that DualAlign can serve as a practical complement to limited gold data and, when scaled, may approach performance that is comparable to models trained on real annotations in certain tasks. Unlike prior synthetic approaches that often lack contextual grounding, DualAlign’s demographic and semantic alignment mechanisms yield training signals that better capture diverse symptom expressions, improving model generalization within the targeted classification tasks.

\begin{table}[h]
\small
\setlength{\tabcolsep}{2.2pt}
\centering
\begin{tabular}{|l|r|r|r|r|}
\hline
\textbf{Category} & \textbf{Gold} & \textbf{Bronze} & \textbf{DualAlign} & \textbf{DualAlign} \\
                 &               &                 & \textbf{(matched)} & \textbf{(full)} \\
\hline
Cognitive impairment      & 6,240  & 2,704  & 3,100  & 82,359 \\
Notice/concern by others  &   785  & 1,710  &   865  & 22,951 \\
Require assistance        & 1,864  & 1,205  &   525  & 13,915 \\
Physiological changes     & 1,340  & 1,769  & 2,480  & 65,943 \\
Neuropsychiatric symptoms & 1,342  & 1,308  & 1,720  & 45,693 \\
\hline
\textbf{Total}            & 11,571 & 8,696  & 8,690  & \textbf{233,014} \\
\hline
\end{tabular}
\vspace{1mm}
\caption{Sentence counts per category across datasets. DualAlign (matched) is subsampled to match bronze scale while preserving category proportions from full DualAlign.}
\label{tab:category_stats}
\end{table}

\begin{table}[t]
\centering
\small
\setlength{\tabcolsep}{2.5pt}  
\renewcommand{\arraystretch}{1.0}  
\begin{tabular}{|l|c|c|c|c|}
\hline
\textbf{Training Set (Positive)} & \textbf{Precision} & \textbf{Recall} & \textbf{F1} & \textbf{Accuracy} \\
\hline
Gold Only & 0.87&  0.61&0.72 &0.87  \\
+Bronze & 0.89 & 0.69 & 0.77 & 0.91 \\
+DualAlign (matched) & 0.90 & 0.72 & 0.79  & 0.93 \\
+DualAlign (full) & \textbf{0.92 } & \textbf{0.77 } & \textbf{0.84 } & \textbf{0.95} \\ \hline
DualAlign (full) only & 0.45 & 0.49 & 0.47 & 0.82 \\
\hline
\end{tabular}
\caption{Binary classification performance on the gold test set, with negative samples selected from real world notes at a 5:1 ratio.}
\label{tab:binary_perf_transposed}
\end{table}

\begin{table}[htbp]
\centering
\scriptsize 
\setlength{\tabcolsep}{6.5 pt} 
\renewcommand{\arraystretch}{1.1}
\begin{tabular}{|l|c|c|c|c|c|c|}
\hline
\textbf{Training Set} & \textbf{Acc.} & \textbf{Cog.} & \textbf{Conc.} & \textbf{Requ.} & \textbf{Phys.} & \textbf{Neuro.} \\
\hline
Gold & 0.70 & 0.77 & 0.46 & 0.57 & 0.62 & 0.71 \\
+Bronze & 0.75 & 0.79 & 0.53 & 0.67 & 0.74 & 0.78 \\
+DualAlign (matched) & 0.77 & 0.81 & 0.54 & 0.70 & 0.77 & 0.81 \\
\textbf{+DualAlign (full)} & \textbf{0.80} & \textbf{0.82} & \textbf{0.60} & \textbf{0.72} & \textbf{0.78} & \textbf{0.86} \\ \hline
DualAlign (full) only & 0.53 & 0.60 & 0.22 & 0.39 & 0.48 & 0.62 \\
\hline
\end{tabular}
\caption{Performance on the multi-class classification task (sentence-level AD signs/symptoms classification). \textbf{Acc.} reports overall accuracy; the remaining columns report F1 scores for each symptom category.}
\label{tab:multiclass_f1}
\end{table}

\subsection{Human Evaluation}

To assess the quality of DualAlign-generated data, we conducted a human evaluation with two clinical experts. They independently reviewed synthetic outputs along two dimensions: (1) sentence-level realism, linguistic diversity, structural and clinical complexity, and labeling accuracy on 100 radomly sampled notes; and (2) patient-level narrative coherence and temporal progression on three complete synthetic patient trajectories. 

Reviewers provided qualitative feedback on aspects such as symptom plausibility, annotation correctness, and consistency in disease unfolding. Their observations are summarized below.

\subsubsection{Sentence-level Evaluation}

Experts noted that DualAlign-generated sentences showed greater contextual richness, specificity, and temporal plausibility compared to prior unconstrained LLM generations (bronze data). However, they identified isolated factual inconsistencies—such as clinical mischaracterizations of anosmia and hyposmia—and inconsistencies in symptom-to-stage mappings, particularly regarding MMSE and MoCA score interpretations. Annotation accuracy was estimated at around 85\%, with higher error rates in categories like \textit{concerns raised by others} and \textit{requires assistance}. Frequent errors included misinterpretation of negation, semantic ambiguity, and overgeneralization of functional difficulties.

\subsubsection{Longitudinal Evaluation}

Reviewers found that while the notes offered strong symptom coverage, the temporal development of disease trajectories was less realistic. In several cases, symptoms appeared overly compressed into single notes, and cognitive transitions (e.g., from MCI to dementia) lacked nuance or consistency. Inconsistencies in clinical staging and timeline compression were noted as areas for improvement. Nonetheless, reviewers emphasized that the structure and breadth of content represented a clear advance over prior synthetic datasets.

\section{Discussion and Limitations}

DualAlign is a novel framework that advances the clinical realism of synthetic EHR data by bridging statistical fidelity and narrative plausibility through two synergistic mechanisms. First, demographic and risk factor–guided prompting conditions generation on patient attributes, including age, sex, race/ethnicity, and SDOH, to simulate realistic and demographically diverse clinical personas. Second, symptom trajectory alignment guides the inclusion of AD-related signs and symptoms based on their empirically observed distributions across disease stages, producing temporally grounded, context-rich narratives.

Our results show that DualAlign substantially advances sentence-level synthetic data generation for Alzheimer’s research. Compared to simple prompt-based LLM generation, it yields more diagnostically informative and diverse symptom descriptions. These gains were validated through both expert review and empirical model performance: DualAlign-enhanced training data consistently outperformed prior synthetic baselines in binary and multi-class classification tasks. Experts further noted significant improvements in contextual specificity, sentence diversity, and temporal plausibility—key qualities for high-utility clinical NLP datasets.

Despite its advances, DualAlign exhibits several limitations. First, longitudinal narrative coherence remains challenging: while the model generates multi-year trajectories, symptom progression is sometimes compressed and transitions between cognitive stages can be abrupt or inconsistent. This reflects known difficulties in LLMs with long-range dependencies and structured temporal reasoning—key for modeling chronic diseases like Alzheimer’s. Future work may benefit from planning-based generation or memory-augmented architectures to improve temporal fidelity. 
Second, annotation accuracy—though guided by structured protocols—showed ~15\% error rate in fine-grained categories (e.g., requires assistance, concerns by others), primarily due to linguistic ambiguity, negation, and overgeneralization. Enhanced semantic parsing or post-hoc correction modules could improve label reliability. 
Third, while DualAlign improves representation of underrepresented subgroups, generated cohorts exhibit residual homogeneity—likely due to mode collapse—and occasionally include rare, clinically implausible risk combinations. Integrating domain-informed constraints or rejection sampling could enhance phenotypic realism. 
Finally, human evaluation, though conducted by two clinicians, was limited in scale (100 notes, 3 longitudinal cases). Future validation should include larger evaluations with more rigorous analysis.

Our code and resources are available via an anonymized GitHub repository. We also release our synthetic notes and the associated AD signs and symptoms dataset. The human-annotated gold standard, derived from the U.S. Department of Veterans Affairs (VA), is accessible through VA approval.

\section{Conclusion}
    DualAlign advances sentence-level synthetic EHR generation for AD by enhancing contextual realism, symptom diversity, and subgroup coverage within a scalable, privacy-preserving framework. Experiments show  gains in text diversity and downstream model performance, representing a promising step toward realistic longitudinal patient simulation.   
    Yet the limitations it exposes—particularly in temporal modeling, annotation precision, and distributional control—highlight that robust synthetic data is necessary but not sufficient for clinical deployment. Closing these gaps will require deeper integration of clinical knowledge and rigorous evaluation standards.   

\bibliography{CameraReady/LaTeX/aaai25}

\onecolumn
\section{Appendix}
\subsection{AD Risk Factors} \label{sec:ad-risk-factors}
\begin{longtable}{l l c}
\multicolumn{1}{c}{\textbf{Factor}} & \textbf{Category} & \textbf{Prevalence (\%)} \\
\midrule
\endhead

\bottomrule
\endfoot

\rowcolor{lightgray} \multicolumn{3}{l}{\textbf{1. DEMOGRAPHIC \& SOCIOECONOMIC FACTORS}} \\
\midrule
Age & $<$65 (early-onset) & 8.0\% \\
Age & 65--74 & 22.0\% \\
Age & 75--84 & 45.0\% \\
Age & $\geq$85 & 25.0\% \\
\addlinespace

Gender & Male & 42.0\% \\
Gender & Female & 56.0\% \\
Gender & Non-binary/Other & 2.0\% \\
\addlinespace

Race & White & 58.0\% \\
Race & Black or African American & 22.0\% \\
Race & Asian & 8.0\% \\
Race & American Indian or Alaska Native & 0.5\% \\
Race & Native Hawaiian or Other Pacific Islander & 0.5\% \\
Race & Mixed/Multiracial & 6.0\% \\
Race & Others/unknown & 5.0\% \\
\addlinespace

Ethnicity & Hispanic/Latino & 15.0\% \\
Ethnicity & Non-Hispanic/Latino & 80.0\% \\
Ethnicity & Others/unknown & 5.0\% \\
\addlinespace

Geographic Location & Urban & 55.0\% \\
Geographic Location & Suburban & 30.0\% \\
Geographic Location & Rural & 15.0\% \\
\addlinespace

Education Level & No formal education & 3.0\% \\
Education Level & Primary & 25.0\% \\
Education Level & Secondary & 45.0\% \\
Education Level & College & 20.0\% \\
Education Level & Postgraduate & 7.0\% \\
\addlinespace

Financial Status & Low income & 38.0\% \\
Financial Status & Middle income & 55.0\% \\
Financial Status & High income & 7.0\% \\
\addlinespace

Employment/Occupation & Retired & 65.0\% \\
Employment/Occupation & Manual labor & 20.0\% \\
Employment/Occupation & Professional & 10.0\% \\
Employment/Occupation & Unemployed & 5.0\% \\
\addlinespace

Health Insurance & None & 5.0\% \\
Health Insurance & Public (e.g., Medicare/Medicaid) & 75.0\% \\
Health Insurance & Private & 20.0\% \\
\addlinespace

Health Literacy & Low & 35.0\% \\
Health Literacy & Moderate & 50.0\% \\
Health Literacy & High & 15.0\% \\
\addlinespace

Housing Instability & Stable & 82.0\% \\
Housing Instability & Unstable (eviction/foreclosure) & 15.0\% \\
Housing Instability & Homeless & 3.0\% \\
\addlinespace

\rowcolor{lightgray} \multicolumn{3}{l}{\textbf{2. MEDICAL \& BIOLOGICAL FACTORS}} \\
\midrule
Family History of AD & Yes & 28.0\% \\
Family History of AD & No & 72.0\% \\
\addlinespace

Hypertension & Yes & 68.0\% \\
Hypertension & No & 32.0\% \\
\addlinespace

Diabetes & Yes & 34.0\% \\
Diabetes & No & 66.0\% \\
\addlinespace

Cardiovascular Disease & Yes & 45.0\% \\
Cardiovascular Disease & No & 55.0\% \\
\addlinespace

Obesity & Yes & 41.0\% \\
Obesity & No & 59.0\% \\
\addlinespace

Stroke History & Yes & 18.0\% \\
Stroke History & No & 82.0\% \\
\addlinespace

Autoimmune Disorders & Yes & 12.0\% \\
Autoimmune Disorders & No & 88.0\% \\
\addlinespace

Traumatic Brain Injury (TBI) & Yes & 9.0\% \\
Traumatic Brain Injury (TBI) & No & 91.0\% \\
\addlinespace

Epilepsy & Yes & 4.0\% \\
Epilepsy & No & 96.0\% \\
\addlinespace

Chronic Inflammation & Yes & 27.0\% \\
Chronic Inflammation & No & 73.0\% \\
\addlinespace

Depression Diagnosis & Diagnosed & 22.0\% \\
Depression Diagnosis & Undiagnosed & 15.0\% \\
Depression Diagnosis & Untreated & 8.0\% \\
\addlinespace

Anxiety Diagnosis & Diagnosed & 18.0\% \\
Anxiety Diagnosis & Undiagnosed & 12.0\% \\
Anxiety Diagnosis & Untreated & 7.0\% \\
\addlinespace

Bipolar Disorder Diagnosis & Diagnosed & 4.0\% \\
Bipolar Disorder Diagnosis & Undiagnosed & 2.0\% \\
Bipolar Disorder Diagnosis & Untreated & 1.0\% \\
\addlinespace

Schizophrenia Diagnosis & Diagnosed & 3.0\% \\
Schizophrenia Diagnosis & Undiagnosed & 1.0\% \\
Schizophrenia Diagnosis & Untreated & 0.5\% \\
\addlinespace

PTSD & Yes & 11.0\% \\
PTSD & No & 89.0\% \\
\addlinespace

Hearing Loss Severity & None & 45.0\% \\
Hearing Loss Severity & Mild & 35.0\% \\
Hearing Loss Severity & Moderate & 15.0\% \\
Hearing Loss Severity & Severe & 5.0\% \\
\addlinespace

Vision Loss Severity & None & 50.0\% \\
Vision Loss Severity & Mild & 30.0\% \\
Vision Loss Severity & Moderate & 15.0\% \\
Vision Loss Severity & Severe & 5.0\% \\
\addlinespace

Chronic Pain & Yes & 39.0\% \\
Chronic Pain & No & 61.0\% \\
\addlinespace

Acute Pain & Yes & 25.0\% \\
Acute Pain & No & 75.0\% \\
\addlinespace

Physical Disability & Yes & 33.0\% \\
Physical Disability & No & 67.0\% \\
\addlinespace

Cognitive Disability & Yes & 28.0\% \\
Cognitive Disability & No & 72.0\% \\
\addlinespace

\rowcolor{lightgray} \multicolumn{3}{l}{\textbf{3. LIFESTYLE \& ENVIRONMENTAL FACTORS}} \\
\midrule
Diet Type & Balanced & 48.0\% \\
Diet Type & Poor (high processed foods) & 52.0\% \\
\addlinespace

Substance Abuse (legal/illicit) & Yes & 17.0\% \\
Substance Abuse (legal/illicit) & No & 83.0\% \\
\addlinespace

Smoking Status & Never & 45.0\% \\
Smoking Status & Former & 35.0\% \\
Smoking Status & Current & 20.0\% \\
\addlinespace

Alcohol Use & None & 40.0\% \\
Alcohol Use & Moderate & 50.0\% \\
Alcohol Use & Heavy & 10.0\% \\
\addlinespace

Physical Activity Level & Sedentary & 55.0\% \\
Physical Activity Level & Moderate & 35.0\% \\
Physical Activity Level & Active & 10.0\% \\
\addlinespace

Sleep Patterns & Regular & 60.0\% \\
Sleep Patterns & Irregular & 40.0\% \\
\addlinespace

Air Pollution Exposure & Yes & 35.0\% \\
Air Pollution Exposure & No & 65.0\% \\
\addlinespace

\rowcolor{lightgray} \multicolumn{3}{l}{\textbf{4. PSYCHOSOCIAL \& STRESS-RELATED FACTORS}} \\
\midrule
Physical Abuse & Yes & 7.0\% \\
Physical Abuse & No & 93.0\% \\
\addlinespace

Emotional Abuse & Yes & 15.0\% \\
Emotional Abuse & No & 85.0\% \\
\addlinespace

Sexual Abuse & Yes & 4.0\% \\
Sexual Abuse & No & 96.0\% \\
\addlinespace

Combat Exposure & Yes & 6.0\% \\
Combat Exposure & No & 94.0\% \\
\addlinespace

Racism/Discrimination & Yes & 22.0\% \\
Racism/Discrimination & No & 78.0\% \\
\addlinespace

Legal Problems & Yes & 9.0\% \\
Legal Problems & No & 91.0\% \\
\addlinespace

Cultural Stigma Around AD & Yes & 31.0\% \\
Cultural Stigma Around AD & No & 69.0\% \\
\addlinespace

Internalized Shame/Guilt & Yes & 19.0\% \\
Internalized Shame/Guilt & No & 81.0\% \\
\addlinespace

Social Engagement & High (regular social interaction) & 35.0\% \\
Social Engagement & Moderate & 45.0\% \\
Social Engagement & Isolated & 20.0\% \\
\addlinespace

Marital Status & Single & 15.0\% \\
Marital Status & Married & 50.0\% \\
Marital Status & Divorced & 25.0\% \\
Marital Status & Widowed & 10.0\% \\
\addlinespace

Caregiver Availability & Family & 65.0\% \\
Caregiver Availability & Professional caregiver & 25.0\% \\
Caregiver Availability & None & 10.0\% \\
\addlinespace

Stress Levels & Low & 25.0\% \\
Stress Levels & Moderate & 50.0\% \\
Stress Levels & High & 25.0\% \\
\addlinespace

\rowcolor{lightgray} \multicolumn{3}{l}{\textbf{5. ACCESS TO CARE \& STRUCTURAL BARRIERS}} \\
\midrule
Proximity to Healthcare & Easy access & 60.0\% \\
Proximity to Healthcare & Limited access & 30.0\% \\
Proximity to Healthcare & Hard & 10.0\% \\
\addlinespace

Public Transport Access & Easy access & 55.0\% \\
Public Transport Access & Limited access & 30.0\% \\
Public Transport Access & No access & 15.0\% \\
\addlinespace

Primary Language & English & 82.0\% \\
Primary Language & Spanish & 12.0\% \\
Primary Language & Other & 6.0\% \\
\addlinespace

\rowcolor{lightgray} \multicolumn{3}{l}{\textbf{6. DEVELOPMENTAL \& LIFECOURSE FACTORS}} \\
\midrule
Childhood Trauma & Yes & 13.0\% \\
Childhood Trauma & No & 87.0\% \\
\addlinespace

Undocumented Immigrant Status & Yes & 4.0\% \\
Undocumented Immigrant Status & No & 96.0\% \\
\addlinespace

\end{longtable}

\subsection{Note Statictics}

\begin{table}[htbp]
\centering
\caption{Average Number of Notes per Patient by Note Type and Time Window Before Diagnosis}
\label{tab:note_types}
\begin{tabular}{lcccc}
\toprule
\textbf{Note Type} & \textbf{10--7 Years} & \textbf{6--4 Years} & \textbf{3--2 Years} & \textbf{1 Year} \\
                   & \textbf{Before}      & \textbf{Before}     & \textbf{Before}     & \textbf{Before} \\
\midrule
Primary care & 2.54 & 2.59 & 2.95 & 5.01 \\
Neurology & 0.31 & 0.74 & 1.18 & 2.51 \\
Memory clinic & 0.31 & 0.74 & 1.18 & 2.51 \\
Neuropsychology & 0.31 & 0.74 & 1.18 & 2.51 \\
Geriatrics & 1.02 & 0.74 & 0.59 & 1.67 \\
Psychiatry/Mental health & 0.51 & 0.74 & 0.89 & 1.67 \\
Emergency visits & 1.02 & 1.11 & 1.18 & 2.51 \\
Home-based primary care (HBPC) & 0.00 & 0.00 & 0.59 & 1.67 \\
\bottomrule
\end{tabular}
\end{table}

\subsection{Keyword Distribution}

\begin{table}[htbp]
\centering
\caption{Average Keyword Count per Note by Years Before Diagnosis}
\label{tab:keyword_count}
\begin{tabular}{|l|l|}
\toprule
\textbf{Years Before Diagnosis} & {\textbf{Avg. Keywords per Note}} \\
\midrule
10 & 2.745 \\
9  & 2.874 \\
8  & 2.993 \\
7  & 3.101 \\
6  & 3.272 \\
5  & 3.384 \\
4  & 3.508 \\
3  & 3.678 \\
2  & 3.829 \\
1  & 4.160 \\
\bottomrule
\end{tabular}
\end{table}

\begin{table}[htbp]
\centering
\caption{Average Keyword Proportion by Category (Relative to Memory = 1)}
\label{tab:keyword_proportion}
\begin{tabular}{l|l}
\toprule
\textbf{Keyword Category} & {\textbf{Proportion}} \\
\midrule
Speech/language & 2.746 \\
Memory & 1.000 \\
Learning/Perception & 1.733 \\
Assistance Needed & 1.531 \\
Physiological Changes & 8.766 \\
Neuropsychiatric Symptoms & 4.399 \\
\bottomrule
\end{tabular}
\end{table}

\subsection{Example Note} \label{sec:example-note}

\begin{tcolorbox}[title={\textbf{Clinical Note}}, colback=white, colframe=blue!75!black, coltitle=black, fonttitle=\bfseries, rounded corners, boxrule=1pt]

\noindent
\textbf{Patient Name}: [redacted] \hfill \textbf{Date}: [redacted] \\
\textbf{Patient ID}: [redacted] \hfill \textbf{DOB}: [redacted] \\
\textbf{Provider}: Dr. [redacted] \hfill \textbf{Visit Type}: Primary Care

\vspace{6pt}

\noindent
\textbf{Subjective}: \\
The patient is a 70-year-old African American female, currently retired, living in urban conditions with unstable housing. She reports moderate levels of stress and is experiencing increased forgetfulness, which she notices mainly in her daily activities, such as misplacing household items and missing appointments. Her daughter, who accompanied her, mentioned the patient has been asking the same questions repeatedly. The patient is also experiencing difficulty with balance and has had minor falls at home. She reports mild vision loss but denies any significant hearing loss (HoH). Though she has no serious auditory issues, she sometimes misses parts of conversations. The patient complains of chronic pain exacerbated by her physical disability. Sleep patterns are irregular, and she describes symptoms consistent with insomnia. The daughter reports some issues with smell (hyposmia) but denies complete anosmia. The patient also notes occasional difficulty swallowing, though no significant dysphagia is observed. Recent challenges with incontinence have been causing distress.

\vspace{6pt}

\noindent
\textbf{Objective}: 
[leftmargin=*,noitemsep]
    
- \textbf{Vital Signs}: BP 150/95 mmHg (hypertensive), HR 82 bpm, Temp 98.6°F
    
- \textbf{Physical Exam}: No acute distress, gait appears slightly unsteady, mild vision loss confirmed. Hearing (auditory) screening shows within normal range, but sporadic minor misses in auditory cues reported.
    
- \textbf{Neurological Examination}: Balance slightly impaired; Mini-Mental State Examination (MMSE) reveals a score of 26/30, indicating possible early cognitive decline.
    
- \textbf{Recent labs}: Elevated cholesterol levels consistent with cardiovascular disease; no hyperglycemia present.
    
- \textbf{Cognitive testing}: Reflects mild cognitive impairment, consistent with early signs of Alzheimer's disease.

\vspace{6pt}

\noindent
\textbf{Assessment}: \\
The patient presents with early prodromal signs that could be consistent with developing Alzheimer's disease, characterized by forgetfulness, imbalance issues, and mild cognitive impairment. Her cardiovascular disease and lifestyle, including poor diet and sedentary activity, could contribute to cognitive decline. Chronic inflammation due to her physical conditions and comorbid depression and bipolar disorder could also be influencing factors. Differential diagnoses include:
\begin{enumerate}
    \item Early Alzheimer's disease
    \item Vascular dementia related to her significant cardiovascular disease
    \item Mild Cognitive Impairment (MCI) secondary to depression and lifestyle factors
\end{enumerate}

\vspace{6pt}

\noindent
\textbf{Plan}:
\begin{enumerate}
    \item Encourage lifestyle modifications, including improved diet and regular physical activity, to address cardiovascular risks and potentially slow cognitive decline.
    \item Referral to a neurologist for further evaluation and management of cognitive decline.
    \item Adjust current antihypertensive regimen to better manage high blood pressure, possibly improving cognitive and balance issues.
    \item Recommend a sleep hygiene program and consider non-pharmacological interventions for insomnia.
    \item Monitor any worsening of dysphagia or balance issues and consider referral to specialists if symptoms persist or worsen.
    \item Counseling and support for depression and bipolar disorder, possibly adjusting psychiatric medications if necessary.
    \item Discuss social support and housing stability solutions to reduce stressors that may exacerbate symptoms.
\end{enumerate}

\vspace{6pt}

\noindent
\textbf{Follow-up}: Scheduled in 3 months for reassessment and ongoing management of symptoms.

\vspace{6pt}

\noindent
\textbf{Provider Signature}: Dr. [redacted] \hfill \textbf{Date}: [redacted]

\end{tcolorbox}
\begin{multicols}{2}
\subsection{Temporal Context}\label{sec:temporalcontext}
Year before AD : AD development stage
            10: "Early prodromal stage", \\
            9: "Early prodromal stage", \\
            8: "Early prodromal stage", \\
            7: "Early prodromal stage", \\
            6: "Mild cognitive impairment stage",\\
            5: "Mild cognitive impairment stage",\\
            4: "Mild dementia stage",\\
            3: "Mild dementia stage",  \\          
            2: "Moderate dementia stage" ,         \\
            1: "Moderate dementia stage"
            \\
\subsection{Keywords for Each Category} \label{sec:keyword-list}
\begin{description}
\item[Speech/language]
    
- communication, speech, speaking
    
- word-finding, word-retrieval, naming, encoding, phonemic
    
- aphasia, paraphasia, anomia, dysnomia
    
- fluency, perseveration, repetition
    
- language, linguistic
    
- comprehend, understand, alexia

\item[Memory]

- memory, amnesia, amnestic
    
- remembering, recognizing, recall, recount, retain
    
- forget, lapse

\item[Learning/Perception]

- attention, concentration, focus
    
- learning, abstraction, problem-solving
    
- executive function, cognitive, neurocognitive, thinking, processing
    
- visuospatial, multidomain, global, agnosia
    
- getting lost, trouble finding, disoriented, confusion
    
- Handwriting deterioration

\item[Assistance Needed]

- ADLs: eating, dressing, grooming, toileting, bathing, mobility
    
- iADLs: cooking, housekeeping, cleaning, laundry, shopping
    
- phone use, computer use
    
- managing medications, managing bills, managing finances
    
- driving, transportation
    
- medical and legal decision-making
    
- healthcare proxy, HPOA, guardian, guardianship
    
- supervision required

\item[Physiological Changes]

- hearing, auditory, SNHL, HoH
    
- vision
    
- smell, anosmia, hyposmia
    
- swallowing, dysphagia
    
- gait, balance
    
- sleep, insomnia
    
- pain
    
- incontinence

\item[Neuropsychiatric Symptoms]

- mood, affect, behavior, apathy
    
- personality
    
- depressed, anhedonia
    
- anxiety, anxious, agitation, hypervigilance, restless, overwhelmed
    
- insight, judgment, impulsive, anosognosia
    
- anger, short-tempered, irritable, aggressive, shouting
    
- erratic, rummaging
    
- wandering
    
- thought disorder
    
- delusion, hallucination, paranoia, psychosis

\end{description}
\subsection{Annotation Guideline} \label{sec:annotation-guideline}

Class 1: Cognitive impairment 

Cognitive impairment is when a person has trouble remembering, learning, concentrating, or making decisions that affect their everyday life, including patient subjective statements as well as Dr. statements.

    Forgetting appointments and dates.
    Forgetting recent conversations and events.
    Having a hard time understanding directions or instructions.
    Losing your sense of direction.
    Losing the ability to organize tasks.
    Becoming more impulsive.
    Memory loss.
    Frequently asking the same question or repeating the same story over and over (perseveration)
    Not recognizing familiar people and places
    Having trouble exercising judgment, such as knowing what to do in an emergency
    Difficulty planning and carrying out tasks, such as following a recipe or keeping track of monthly bills
    meaningless repetition of own words, lack of restraint, wandering and getting lost
    lose your train of thought or the thread of conversations
    trouble finding your way around familiar environments
    problems with speech or language 
    mental decline, difficulty thinking and understanding, confusion in the evening hours, disorientation, lack of orientation, forgetfulness, making things up, mental confusion, difficulty concentrating, inability to create new memories, inability to do simple math, or inability to recognize common things, poor judgment, impaired communication, poor concentration, difficulty remembering recent conversations, names or events
    forget things more often, forget important events such as appointments or social engagements, issues with recall, changes in abstract reasoning ability

Class 2: Notice/concern by others (not by provider, refers to friends/family and neighbors)

family complains of X (may be related to any class including cognitive decline, functional impairment, Neuropsychiatric and physiological changes)

    noticed changes in ability, speed
    concern expressed by family/friends
    complaints of patient easily angered
    Daughter reports that she repeatedly asks the same question...had difficulties using her smartphone.
    Daughter reports that she has issues with banking...some decrease in personal hygiene, forgets to take meds, forgets where food is in the house, etc.
    Patient has gone out at 1:30 a.m. without telling anyone; they are concerned, but patient always has a response.
    She (daughter) tells me that her mom has repeatedly changed the medications in the pill boxes that she has arranged for her.

Class 3: Requires assistance (defined as from a person)/ functional impairment

    needs help with or loss of ability with ADLs/iADLs, difficulty with self-care, trouble managing belongings
    ADLs: dressing, eating, toileting, bathing, grooming, mobility
    iADLs: housekeeping‐related activities (cleaning, cooking, and laundry) and complex activities (telephone use, medication intake, use of transportation/driving, budget/finance management, and shopping)
    The patient will continue to require assistance with all complex medical, legal and financial decision making.
    She will need 24-hour supervision for her safety.
    Direct supervision is required for medications using a pillbox.
    Best not to have him use stove.
    If left alone for period of time, will need guardian alert or consider camera surveillance.
    He is able to make a meal, to dress himself, to bathe, to shave, but continues to need help with finances.
    Wife has to remind him about appointments, in particular.
    Driving should not be permitted, and he will need assistance with IADLs and decision making.
    Veteran does need assistance with all IADLs and most ADLs.
    Traveling out of neighborhood, driving, arranging to take buses-limited night driving now
    Resides in assisted living facility or nursing home 

    Writing checks, paying bills, balancing checkbook-minimal (automatic payment) N/A
    Playing a game of skill-no hobbies N/A

Class 4: Physiological changes 

    senses: vision, hearing, smell loss, SNHL: sensorineural hearing loss, HoH,
    sleep: Excessive daytime sleepiness, changes in sleep patterns
    speech/swallowing (speech difficulties also in "Cognitive Impairment" class)
    movement/gait/balance
    inability to combine muscle movements: jumbled speech, difficulty speaking, aphasia, dysphasia, difficulty swallowing, dysphagia, difficulty walking, mobility, problems with gait and balance, gait slowing

(Focus on early possible associations, late physiological changes associated with AD like seizure, incontinence etc. are skipped.)

Class 5: Neuropsychiatric symptoms

    mood changes: depression, irritability, aggression, anxiety, apathy, personality changes, behavioral changes, agitation, short-temper, quick to anger, anger issues, mood instability, hypervigilance, negative cognitions, mood lability, labile mood, anhedonia
    feeling increasingly overwhelmed by making decisions and plans
    paranoia, delusions, hallucinations, psychosis, hear voices, see X

\end{multicols}

\end{document}